\documentclass[11pt,a4paper]{article}
\usepackage[hyperref]{naaclhlt2019}
\usepackage{times}
\usepackage{latexsym}
\usepackage{graphicx}
\usepackage{url}
\usepackage{float}
\usepackage{subfigure,amssymb}
\aclfinalcopy 

\makeatletter
\newcommand\thankssymb[1]{\textsuperscript{\@fnsymbol{#1}}}
\makeatother

\pdfoutput=1

\title{Detection and Classification of mental illnesses on social media using RoBERTa}

\author{Ankit Murarka\thanks{\thankssymb{1} These authors contributed equally} \\
  IBM Corporation \\
  Research Triangle Park, NC \\
  {\tt ankit.murarka1} \\
  {\tt@ibm.com} \And
  Balaji Radhakrishnan\thankssymb{1} \\
  {\tt balag59@gmail.com} \\\And
  Sushma Ravichandran\thankssymb{1} \\
 IBM Research \\
 Yorktown Heights, NY \\
  {\tt sushma.ravichandran} \\
  {\tt@ibm.com} \\}
\date{}

\begin{document}
\maketitle
\begin{abstract}
Given the current social distancing regulations across the world, social media has become the primary mode of communication for most people. This has resulted in the isolation of many people suffering from mental illnesses who are unable to receive assistance in person. They have increasingly turned to social media to express themselves and to look for guidance in dealing with their illnesses. Keeping this in mind, we propose a solution to detect and classify mental illness posts on social media thereby enabling users to seek appropriate help. 
In this work, we detect and classify five prominent kinds of mental illnesses- depression, anxiety, bipolar disorder, ADHD and PTSD by analyzing unstructured user data on social media platforms. In addition, we are sharing a new high-quality dataset to drive research on this topic.
We believe that our work is the first multi-class model that uses a Transformer \citet{NIPS2017_7181}-based architecture such as RoBERTa \citet{roberta} to analyze people's emotions and psychology. We also demonstrate how we stress-test our model using behavioral testing. With this research, we hope to be able to contribute to the public health system by automating some of the detection and classification process.
\end{abstract}
\section{Introduction}
During these unprecedented times when the world is plagued by COVID19, a large number of people have been showing symptoms of clinical anxiety or depression\footnote{https://afsp.org}. This can be attributed to a myriad of reasons including, but not limited to, lock down, mandatory social distancing, higher unemployment, economic depression and work-related stress. 

In a report published earlier this year, the American Foundation for Suicide Prevention found that people experience anxiety (53\%) and sadness (51\%) more often now than before the coronavirus pandemic. Additionally, as per the report, despite physical distancing guidelines, Americans socialize with friends and families at a similar frequency as they used to before the lockdown orders confirming that physical distancing and isolation can contribute to increase in anxiety, stress or depression.

In the past decade, social media has transformed how people interact with each other. Apart from sharing factual information and news, people actively partake in sharing their day to day activities, experiences, feelings, opinions, hopes, desires, and emotions online. These texts provide information which can be used to identify the mental health individuals. Furthermore, the current state of enforced social distancing and isolation has propelled more people to express their emotions on social media as it provides them with an accessible platform to share their thoughts with others, many a times, in search for help. 

Our research work utilizes user data, especially the kind pertaining to emotions as this class of data can give us valuable insights about the mental state of a person; and, in turn, our work has the potential to assist in the diagnosis and analysis of various mental disorders. This study aims to bridge the gap between people in search of help and experts who can provide the needed help.

Due to the paucity of adequate annotated and structured user data in this domain, we decided to generate our own dataset by crawling subreddits on reddit.com\footnote{https://www.reddit.com} pertaining to our use case as a lot of users were found to have shared their feelings there. Although Reddit was the sole source of our dataset, we believe that this study can be seamlessly extended to other social media platforms as well because of the presence of similar unstructured user data online.

The advent of transformers and BERT \citet{bert:2018} has caused quite a stir in the NLP community because of the state-of-the-art results it was able to produce in various NLP tasks. In this work, we use a RoBERTa \citet{roberta} based classifier, which has a similar architecture to BERT with an improved pre-training procedure. RoBERTa's effective and efficient performance on unstructured data and its ability to learn contextual information compelled us to explore and capitalize its power to categorize online user generated texts into various classes of Mental Illness. 

We identify five broad classes of mental illnesses - depression, anxiety, bipolar disorder, ADHD (Attention Deficit Hyperactivity Disorder), PTSD (Post Traumatic Stress Disorder) and an additional 'None' class (which does not pertain to any mental illness). We train a multi-class classifier on the data crawled from online user data. Based on our experiments, we present encouraging results that demonstrate that social media data has the potential to complement standard clinical procedures in the prognosis of mental health amongst two broad categories of users - ones who are seeking help online and ones who are unbeknownst of their condition.

\section{Related Work}
In the recent past, people have increasingly turned to social media to share and seek counsel on the topic of mental health. This has prompted researchers to utilize the information and apply a plethora of techniques in NLP and Machine Learning in order to assist people who might require help. Most of the recent research has revolved around Reddit data as in the case of \citet{kim2020deep}, \citet{Gkotsis2017mentalhealth}, \citet{sekulic-strube-2019-adapting}, \citet{zirikly:2019}. Prior to this recent shift to Reddit data, a lot of the earlier research was focused on utilizing Twitter data as in the case of \citet{Orabi:2018}, \citet{benton2017multitask}, \citet{Coppersmith2015Shared}. 

There have been a wide variety of approaches ranging from classical NLP techniques to neural network based deep learning methods. \citet{Coppersmith2015Shared} used character level language models to examine how likely a sequence of characters is to be generated by a user with mental health issues. \citet{benton2017multitask} evaluated a standard regression model, a multilayer perceptron single-task learning (STL) model, and a neural MTL model on detecting multiple types of mental health issues. \citet{Orabi:2018} utilized word embeddings in tandem with a variety of neural network models like CNNs and RNNs to detect depression. \citet{Gkotsis2017mentalhealth} experimented with Feed Forward Neural Networks, CNNs, SVMs and Linear classifiers to perform binary classification on mental health posts. \citet{sekulic-strube-2019-adapting} came up with the approach of using Hierarchical Attention Networks(HANs) to detect a wide range of mental health issues like Depression, ADHD, Anxiety etc. and trained a binary classifier for each of the disorders. The most recent work on this was by \citet{kim2020deep} who proposed a CNN-based classification model. Once again though, each disorder had its own separate binary classifier to perform the detection. 

To our knowledge, this is the first attempt at treating this problem as a multi-class classification problem, where a single classifier can not only detect the disorder, but, also accurately classify the type of disorder that the person is referring to in their post. In addition, this is also the first work that harnesses the incredible capabilities of an advanced Transformer based algorithm like RoBERTa to solve this difficult problem.

\section{Dataset}
The Reddit API was used to crawl 13 subreddits and 17159 posts text and title text to obtain data for this research work. The text from comment threads was not collected as it tended to diverge from the main topic of the subreddit. Out of the 13 subreddits, 5 can be directly associated with a mental illness. They are: {\small\verb|bipolar|}, {\small\verb|adhd|}, {\small\verb|anxiety|}, {\small\verb|depression|} and {\small\verb|ptsd|}. The posts in these subreddits were assigned a class label corresponding to the name of the subreddits. The remaining 8 subreddits were chosen from a wide range of topics. They are {\small\verb|music|}, {\small\verb|travel|}, {\small\verb|india|}, {\small\verb|politics|}, {\small\verb|english|}, {\small\verb|datasets|}, {\small\verb|mathematics|} and {\small\verb|science|}. These general topic subreddits were combined together and assigned the class label {\small\verb|none|}. 

While collecting data, we ensured that the number of upvotes for each post in all subreddits is more than 10. We also set a minimum post token length of 30 tokens. This is so that we retain quality in the dataset. We approximately crawled about 3000 posts under the subreddits dealing with mental illness and about 300 posts for each of the general topic subreddits. This also ensured a good balance of class labels. While selecting the eight general topic subreddits, we not only selected subreddits that have sufficiently high number of posts, but also ensured that we cover a broad range of topics. Table \ref{data-stats} shows the statistics collected for each subreddit. The dataset was preprocessed to remove any URLs or usernames that could potentially contain sensitive information. This was done keeping in mind that the dataset will be released publicly for the purpose of extending this research work.

To gauge the data quality we ran some analysis. We manually went over the lowest voted posts for each mental illness subreddit. We wanted to establish that texts from these posts expressed emotions from people discussing corresponding mental illness it is labelled as. Table \ref{posts-excerpts} presents excerpts of lowest voted post from each mental illness subreddit.

We also certified that the general topic subreddits did not have a high similarity with the posts corresponding to other the other 5 subreddits. This was done to ensure that we do not have any false negatives while assigning truth labels. We counted the number of posts the mental illness terms appeared in, for each subreddit. The subreddits corresponding to mental illnesses had a much higher count of these words. In addition to this, we compared the cosine similarity between some of the highest/lowest posts of mental illness subreddits and the general topic subreddits and manually compared the results to find that this distance was higher than the distance between two posts of the mental illness subreddits.

We also attempted to augment the data using Easy Data Augmentation \citet{eda:2019} to boost the performance of our model. However, we did not observe an apparent shift in our evaluation metrics- explained by the fact the EDA is meant to perform best for smaller datasets ($\leq$ 5000 sample sizes). So, we decided to not advance in that route.

\begin{table*}[!t]
\begin{center}
\begin{tabular}{|c|p{1.5cm}|p{2cm}|p{2cm}|p{1.5cm}|p{1.5cm}|p{1.5cm}|}
\hline
\bf Subreddit & \bf Number of posts & \bf Average no. of words (posts) & \bf Average no. of words (titles) & \bf Average Upvotes & \bf Highest Upvotes & \bf Lowest Upvotes\\ 
\hline
r/depression & 3062 & 152.74 & 12.20 & 517.19 & 4802 & 11\\
r/anxiety & 3027 & 170.38 & 11.75 & 246.07 & 3349 & 11\\
r/ptsd & 2501 & 233.55 & 10.14 & 38.4 & 443 & 11\\
r/adhd & 3082 & 198.55 & 13.71 & 377.13 & 4484 & 11\\
r/bipolar & 3009 & 203.28 & 9.26 & 32.37 & 363 & 11\\
none & 2478 & 238.52 & 15.76 & 6715.33 & 199295 & 11\\
\hline
\end{tabular}
\end{center}
\caption{\label{data-stats} Dataset: Statistics }
\end{table*}

\begin{table}[!t]
\begin{center}
\begin{tabular}{|p{7cm}|}
\hline Lowest rated post \\ \hline
\textbf{r/depression}- The older I am getting the less hope I have to secure a life worth living. I feel finished because I always had that state of mind \\
\textbf{r/adhd}- Does anyone else feel like u do have a personality and ability to make friends but ure kind of stuck in ur own body \\
\textbf{r/bipolar}- I just made a really impulsive choice with my breed of dog because I had to have one NOW. I was thinking about it constantly day and night and I couldn't sleep. \\
\textbf{r/anxiety}- I find myself constantly remembering embarrassing or cringey moments from my past (ranging anywhere from present day to back about 10 years) and cringing hard at them \\
\textbf{r/ptsd}- I feel like I am just constantly angry. Angry about my trauma and how it has affected me, and angry about where I am in my life because of it. I don't want to be angry anymore \\
\hline
\end{tabular}
\end{center}
\caption{\label{posts-excerpts} Dataset: Posts excerpts }
\end{table}

\section{Model}
In this section, we describe our model architecture for the multi-class mental illness classification task. We propose a RoBERTa based classifier in order to accomplish this. In addition, we also compare the proposed model against an LSTM \citet{lstm} based classifier and a BERT \citet{bert:2018} based classifier to demonstrate the superiority of our approach. Since this is an entirely new dataset, there is no established baseline, so the LSTM model will serve as the baseline for our experiments. We also showcase our gains over BERT, the most widely used transformer model today for text classification. 
All our models were implented in Pytorch \citet{NEURIPS2019_9015}. The Transformer models were implemented with the help of the HuggingFace Transformers \citet{Wolf2019HuggingFacesTS} library.

\subsection{LSTM based classifier}
LSTMs(Long Short-Term Memory) were the state of the art models when it came to text classification before the advent of Transformers. They will serve as our baseline. First we tokenized the sentences using NLTK\footnote{https://www.nltk.org/} and converted them to lower case to create our vocabulary. In order to get rid of words that might not exist, we removed all words from our vocabulary that appear only once. We also added {\small\verb|padding|} and {\small\verb|unknown|} to our vocabulary in order to account for padding and unknown tokens respectively. Each sentence was represented using a sequence of length 512 and this forms our input to the LSTM model. We used a 2 layer LSTM for all our experiments with an embedding layer of size 100 and a hidden layer size of 256. Dropout \citet{10.5555/2627435.2670313} with a probability of 0.5 was used in order to achieve regularization. We used standard cross-entropy loss as the loss function. During training, Adam \citet{kingma2014adam} was the optimizer of choice, with a learning rate of 0.005. The model was trained for a total of 25 epochs with a batch size of 32. Gradient clipping was used to prevent exploding gradients.
\subsection{BERT based classifier}
BERT(Bidirectional Encoder Representations from Transformers) has been the biggest breakthrough in the NLP domain in the recent past with state of the art results in a myriad of NLP tasks. Since its inception, better models with gains have been trickling along, but BERT continues to be the most popular model for text classification even today. The BERT classifier comprises a fine-tuned BERT model followed by a dropout layer and a fully connected layer. We fine-tuned a pre-trained BERT-base model on our dataset for this task. A pre-trained tokenizer on BERT is used to tokenize our input sentences. After carefully examining the sentence length distribution, we chose a sequence length of 35 for titles, and 512 for posts and posts+titles. Either padding or truncation was used to ensure that all sentences were represented using the same sequence length. All the BERT based models were fine-tuned on our data for 10 epochs with a learning rate of 1e-5. Adam served as the optimizer and cross-entropy loss was the loss function of choice. A dropout layer with probability of 0.3 was used for the sake of regularization.

\subsection{RoBERTa based classifier}
RoBERTa(Robustly Optimized BERT Pretraining Approach) is another state of the art language model that builds on BERT by modifying key hyperparameters and training on more data. It outperforms BERT on several benchmark tasks and forms the core of our proposed solution. In order to make it a fair comparison with BERT, we retain the architecture and all design choices made with the BERT based classifier barring the pretrained model and the tokenizer which are now all based on RoBERTa. The input sentences were tokenized using a pre-trained tokenizer on RoBERTa-base. Just as in the case of BERT, we chose a sequence length of 35 for titles, and 512 for posts and posts+titles. Similar to BERT, the RoBERTa based models were also fine-tuned for 10 epochs with a learning rate of 1e-5 and Adam. A batch size of 32 was used while fine-tuning on the titles whereas a batch size of 16 was the only viable option to fine-tune on posts and posts+titles. Cross-entropy remained the preferred loss function. Once again, a dropout layer with probability of 0.3 was used for regularization.

\section{Result Analysis}

\begin{table*}
\begin{center}
\begin{tabular}{|l|llll|llll|llll|}
\hline
\bf Models &
\multicolumn{4}{c|}{\bf posts} &
\multicolumn{4}{c|}{\bf titles} &
\multicolumn{4}{c|}{\bf posts+titles} \\ \hline
& P & R & F1 & Acc & P & R & F1 & Acc & P & R & F1 & Acc \\ 
LSTM & 0.74 & 0.72 & 0.72 & 0.72 & 0.65 & 0.64 & 0.64 & 0.64 & 0.77 & 0.76 & 0.76 & 0.76 \\
BERT & 0.83 & 0.82 & 0.82 & 0.82 & 0.72 & 0.71 & 0.71 & 0.71 & 0.87 & 0.87 & 0.87 & 0.87 \\
RoBERTa & 0.86 & 0.86 & \bf 0.86 & 0.86 & 0.73 & 0.72 & \bf 0.72 & 0.72 & 0.89 & 0.89 & \bf 0.89 & 0.89 \\
\hline
\end{tabular}
\end{center}
\caption{\label{res-class} Results: Classification Report }
\end{table*}

\begin{figure*}
\subfigure[Input: posts]{\includegraphics[scale=0.18]{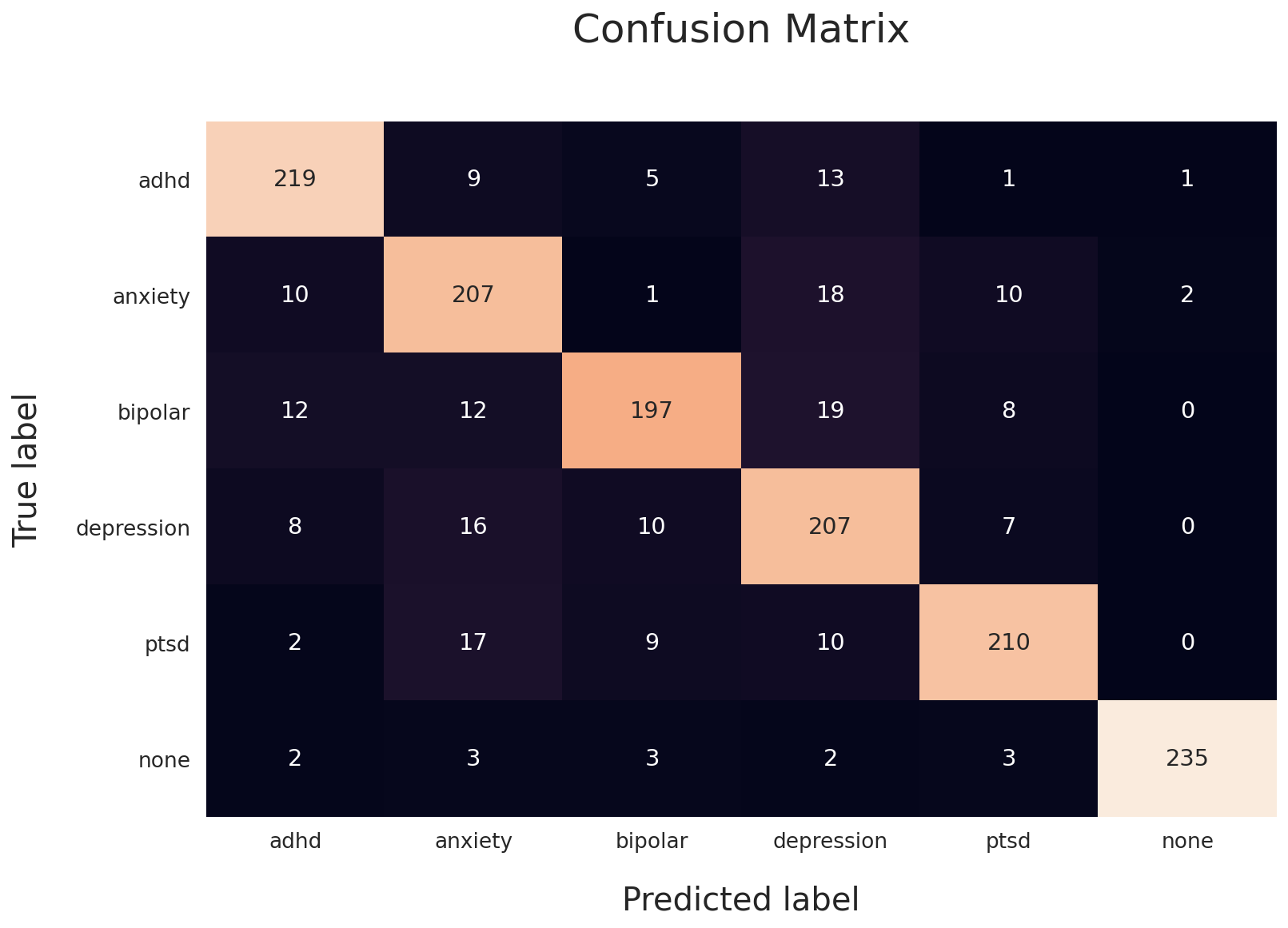}}
\subfigure[Input: titles]{\includegraphics[scale=0.18]{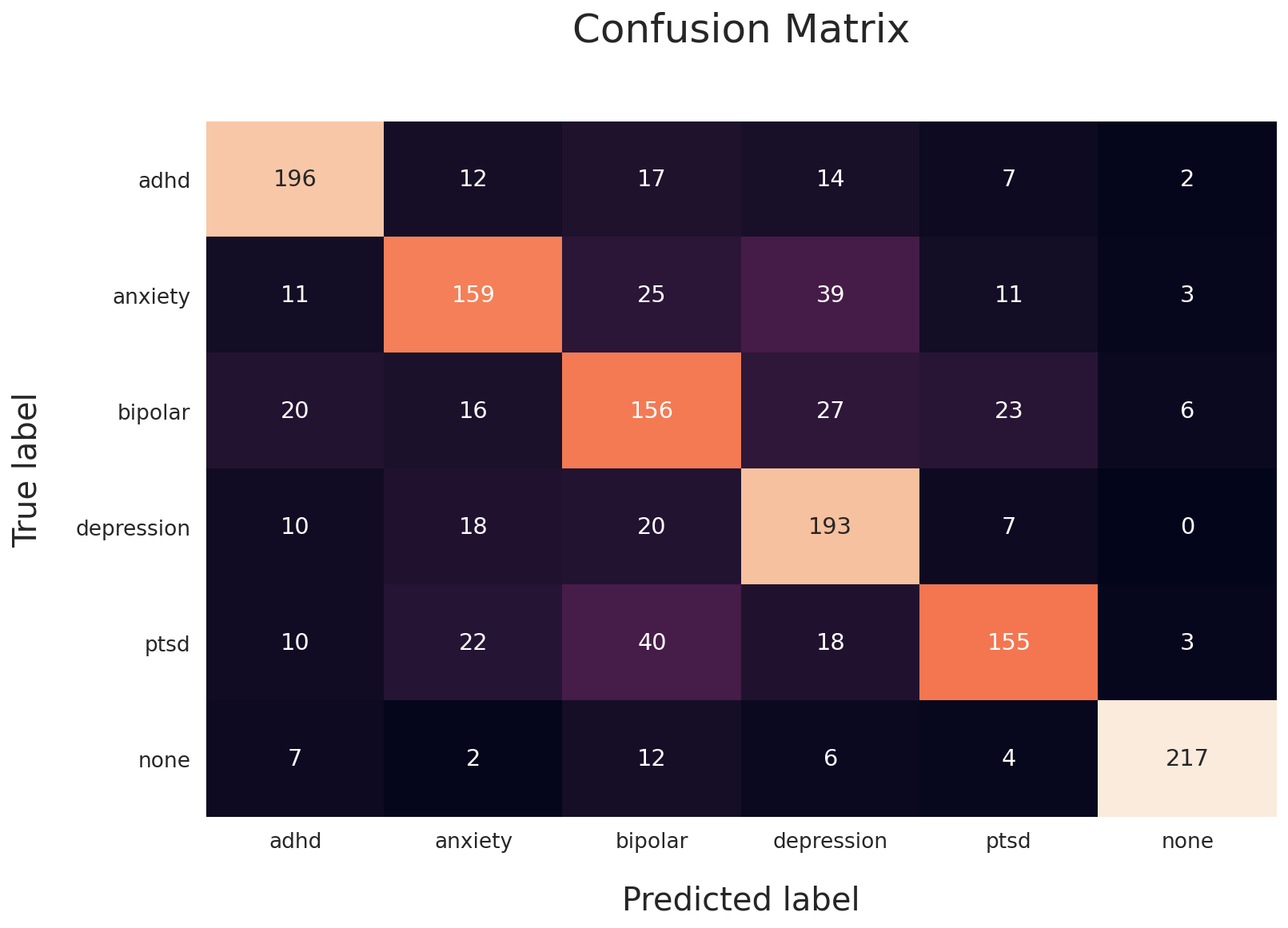}}
\subfigure[Input: posts+titles]{\includegraphics[scale=0.18]{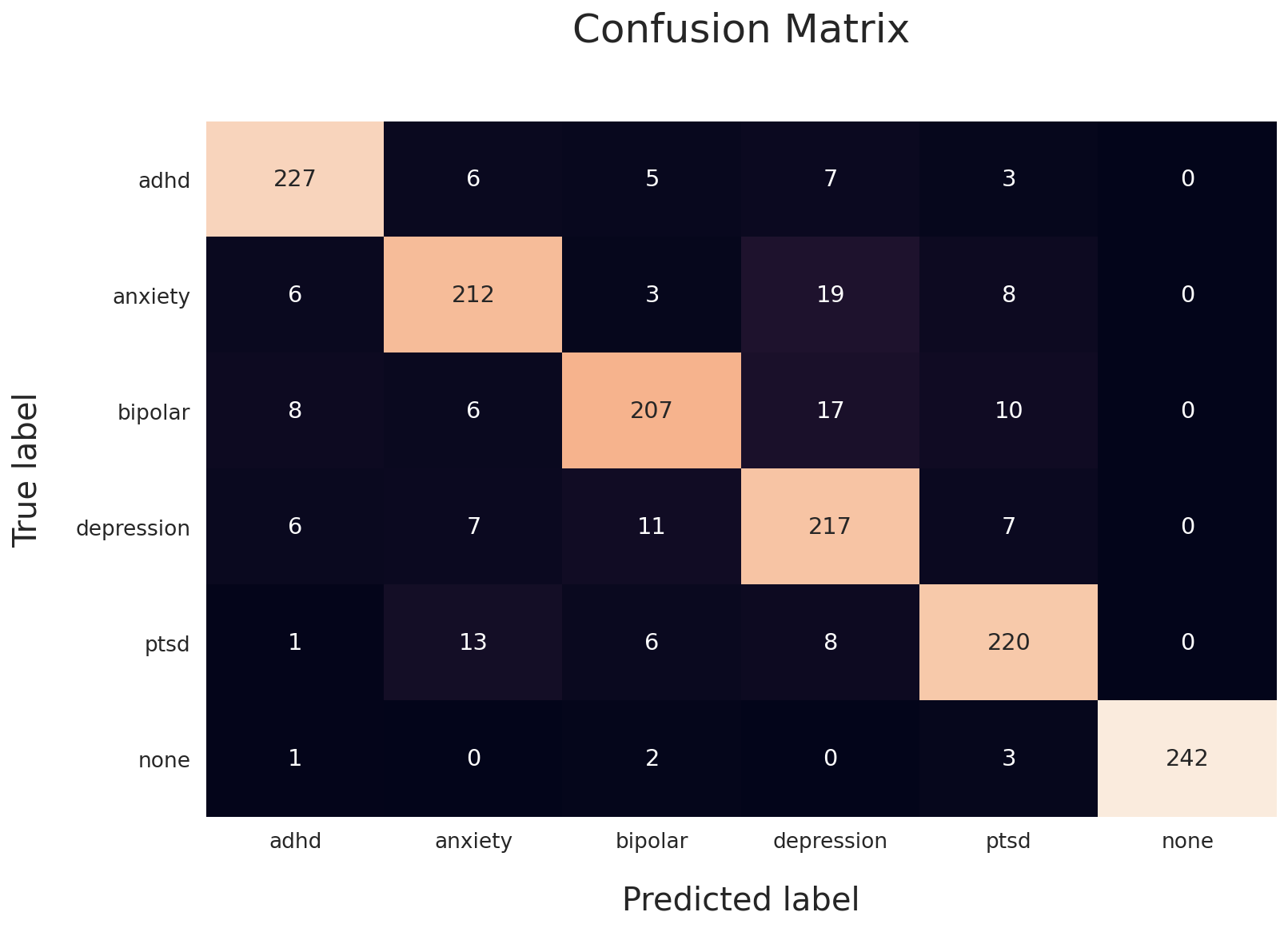}}
\caption{\label{rob-cm}RoBERTa: Confusion Matrices}
\end{figure*}




\begin{table*}[!t]
\begin{center}
\begin{tabular}{|l|lll|lll|lll|}
\hline
\bf Class & \multicolumn{3}{c|}{\bf posts} &\multicolumn{3}{c|}{\bf titles} & \multicolumn{3}{c|}{\bf posts+titles}\\ \hline
& P & R & F1 & P & R & F1 & P & R & F1 \\ 
adhd & 0.87 & 0.88 & 0.87 & 0.77 & 0.79 & 0.78 & 0.91 & 0.92 & 0.91\\
anxiety & 0.78 & 0.83 & 0.81 & 0.69 & 0.64 & 0.67 & 0.87 & 0.85 & 0.86\\
bipolar & 0.88 & 0.79 & 0.83 & 0.58 & 0.63 & 0.60 & 0.88 & 0.83 & 0.86\\
depression & 0.77 & 0.83 & 0.80  & 0.65 & 0.78 & 0.71 & 0.81 & 0.88 & 0.84 \\
ptsd & 0.88 & 0.85 & 0.86 & 0.75 & 0.62 & 0.68 & 0.88 & 0.89 & 0.88\\
none & 0.99 & 0.95 & 0.97 & 0.94 & 0.88 & 0.91 & 1.00 & 0.98 & 0.99 \\
\hline
\end{tabular}
\end{center}
\caption{\label{rob-class} Results: RoBERTa Class-wise results }
\end{table*}


\begin{table*}
\begin{center}
\begin{tabular}{|p{11.5cm}|l|l|l|}
\hline \bf Input & \bf Actual & \bf Predicted \\ \hline
often times i'll get distracted from my thoughts either by external influences or just another idea coming in, and then i have to spend a good 5 minutes trying to work out what i was thinking about again. & adhd & adhd \\
once i come down from flashbacks or panic attacks, i get really bad disassociation. sometimes lasting for days. does anyone else go through this. any tips on how to stop it. i tried grounding but im so far gone it doesn't help. & ptsd & ptsd\\
i can't sit still when i get my eyebrows done, and when i'm in class i usually doodle to focus. i pay attention very well in school regardless of that, and drawing helps me focus. & anxiety & adhd\\
i'm flying from dallas to hong kong in january and it's 17 hours. i've flown 12-13 hour flights before and they really mess with me. so i'm wondering - what are your tips for not going crazy on such a long flight? ps: i'm terrible at sleeping on planes. thinking about taking some sleepy meds to see if it'll help & none & anxiety\\
\hline
\end{tabular}
\end{center}
\caption{\label{test-exa} Results: Interesting Examples }
\end{table*}

As described earlier, in addition to our primary RoBERTa classifier, we also run experiments on an LSTM classifier and a BERT classifier for the sake of comparison. We fine-tune each of the aforementioned models on just the titles, just the posts and a combination of both in order to perform comprehensive tests and comparisons. When combining the titles and posts for our Transformer models, we convert the problem into a sequence-pair classification task. This allows the model to give more importance to the title which would otherwise be lost when combining the title and the post into one single input given the relative difference in their lengths (the average number of tokens in titles is roughly 6\% that of posts).

The results from our experiments are documented in Tables \ref{res-class} through \ref{test-exa}.

As can be observed from Table \ref{res-class}, our proposed RoBERTa based classifier far outperforms the baseline LSTM in all categories. The BERT classifier has results which are quite close to that of RoBERTa's and both beat LSTM by a significant margin, showcasing the incredible capabilities of pre-trained Transformer based architectures. In fact, our RoBERTa model fine-tuned on just the titles was able match the performance of the LSTM model trained on posts. The RoBERTa model was able to achieve an F1 score of 0.86 on the posts and 0.89 on posts+titles which are extremely promising given the complex nature of the multi-class mental illness classification task. The jump in accuracy between posts and posts+titles is not as drastic as the jump between titles and posts. This indicates that the posts offer far more valuable information when compared to the titles and also the fact that most of the useful and relevant information can be extracted from the posts alone. This strong performance on just the posts bodes well for the extensibility of our approach as this can be applied on almost any given social media post without the need for structure in the data like titles, user names, user history, etc. 

The rest of this section will focus solely on the results of our best performing RoBERTa model. Table \ref{rob-class} showcases the granular class-wise results of the RoBERTa model. This table in conjunction with the confusion matrices from Figure \ref{rob-cm} offers us a wealth of useful and interpretable information. 

The first strikingly obvious result is the high accuracy with which the model is able to detect non-illness related posts. Even with just the titles, the model is able to classify the {\small\verb|none|} class with an f1 score of more than 0.9. This gives us hope that this model will suffer from very few false positives when it comes to mental illness detection on social media. 

An even more crucial property of our model can be noticed in the confusion matrices for posts and posts+titles in Figure \ref{rob-cm}. When using posts, just 3 illness related posts across the entire test dataset were misclassified as non-illness posts. This number further reduces to 0 when using titles+posts. This shows that the model will detect mental illness posts correctly nearly every single time, thus ensuring that posts from people who are seeking help never go unnoticed when this solution is deployed in the real world.

When it comes to the class wise performance amongst the mental illnesses, the two best performing classes are {\small\verb|adhd|} and {\small\verb|ptsd|} whereas the two worst performing classes are {\small\verb|depression|} and {\small\verb|anxiety|}. 

The performance of {\small\verb|depression|} and {\small\verb|anxiety|} classes can be attributed to a few factors. The average number of words per post for {\small\verb|depression|} and {\small\verb|anxiety|} are the least for any given class. For instance, {\small\verb|depression|} posts have roughly 53\% lesser textual data when compared to {\small\verb|ptsd|} posts. In addition, studies show that depression might often occur in tandem with another mental illness and our data and results back this up as well. The {\small\verb|depression|} word occurs in 12\% of {\small\verb|anxiety|} posts, 12\% of {\small\verb|ptsd|} posts and 31\% of {\small\verb|bipolar|} posts. Similarly, {\small\verb|anxiety|} occurs in 20\% of {\small\verb|ptsd|} posts, 12\% of {\small\verb|adhd|} posts and 14\% of {\small\verb|bipolar|} posts. This implies that the model cannot give high importance to the mention of these class names like it can with rest of the illnesses, thus making the classification of these 2 classes that much harder.

This can also explain the relatively lower precision scores(higher number of False Positives) for {\small\verb|depression|} and {\small\verb|anxiety|}. When the other illnesses(excluding {\small\verb|depression|} and {\small\verb|anxiety|}) are misclassified, they are almost always misclassified as either {\small\verb|depression|} or {\small\verb|anxiety|}, as can be viewed in Figure \ref{rob-cm}. In the same figure, we can see that {\small\verb|depression|} and {\small\verb|anxiety|} are often misclassified as each other due to the reason that they commonly occur together.

There are more posts in the {\small\verb|adhd|} and {\small\verb|ptsd|} classes that mention the words {\small\verb|depression|} and {\small\verb|anxiety|} than their respective class names itself. One would assume that this would result in subpar results, but, these classes actually perform the best. This really showcases the true potential of our model, where it doesn't just rely on mention of class names, but has a strong understanding of the context of the post itself. Additionally, the symptoms or descriptions provided for these classes could be strong, unique and discriminative enough for the model to be able to classify them correctly even with all the mentions of other class names.

In Table \ref{test-exa} we have documented a few interesting results we observed in the test set. In the first two examples on the table, the RoBERTa model was able to classify the posts correctly without the presence of class names in the input. The prediction is based purely on contextual information learnt about the class labels during the training process. The next two results are interesting because the truth label assigned to the input text may or may not correspond to actual mental illness described in the text. Since, we are not domain experts ourselves, we would need expert intervention to substantiate this theory. As a part of future work, getting professionals to annotate our dataset might help strengthen the model for such examples.

\section{Behavioral Testing}

\begin{table*}
\begin{center}
\begin{tabular}{|p{3cm}|llll|p{3cm}|llll|}
\hline
\multicolumn{5}{|c|}{\bf Synonym Replacement} & \multicolumn{5}{c|}{\bf Label Removal}\\ 
\hline
\bf Test Set Modified & \multicolumn{4}{c|}{\bf posts} & \bf Test Set Modified & \multicolumn{4}{c|}{\bf posts}\\
\hline
 & P & R & F1 & Acc & & P & R & F1 & Acc \\
10\% & 0.86 & 0.85 & 0.85 & 0.85 & 10\% & 0.85 & 0.84 & 0.84 & 0.84\\
50\% & 0.85 & 0.84 & 0.84 & 0.84 & 50\% & 0.81 & 0.80 & 0.80 & 0.80\\
100\% & 0.83 & 0.83 & 0.83 & 0.83 & 100\% & 0.75 & 0.74 & 0.75 & 0.74\\
\hline
\bf Test Set Modified & \multicolumn{4}{c|}{\bf titles} & \bf Test Set Modified & \multicolumn{4}{c|}{\bf titles}\\
\hline
 & P & R & F1 & Acc &  & P & R & F1 & Acc \\
10\% & 0.73 & 0.72 & 0.72 & 0.72 & 10\% & 0.72 & 0.71 & 0.71 & 0.71\\
50\% & 0.71 & 0.71 & 0.71 & 0.71 & 50\% & 0.67 & 0.67 & 0.67 & 0.67\\
100\% & 0.68 & 0.67 & 0.67 & 0.67 & 100\% & 0.61 & 0.61 & 0.60 & 0.61\\
\hline
\multicolumn{5}{|c|}{\bf Label Replace: 'illness'} & \multicolumn{5}{|c|}{\bf Label Replace: random}\\ 
\hline
\bf Test Set Modified & \multicolumn{4}{c|}{\bf posts} & \bf Test Set Modified & \multicolumn{4}{c|}{\bf posts}\\
\hline
 & P & R & F1 & Acc &  & P & R & F1 & Acc\\
10\% & 0.84 & 0.83 & 0.84 & 0.83 & 10\% & 0.83 & 0.82 & 0.83 & 0.8\\
50\% & 0.78 & 0.77 & 0.77 & 0.77 & 50\% & 0.71 & 0.71 & 0.71 & 0.71\\
100\% & 0.70 & 0.67 & 0.68 & 0.67 & 100\% & 0.58 & 0.57 & 0.57 & 0.57\\
\hline
\bf Test Set Modified & \multicolumn{4}{c|}{\bf titles} & \bf Test Set Modified & \multicolumn{4}{c|}{\bf titles}\\
\hline
 & P & R & F1 & Acc &  & P & R & F1 & Acc \\
10\% & 0.72 & 0.71 & 0.71 & 0.71 & 10\% & 0.71 & 0.71 & 0.71 & 0.71\\
50\% & 0.67 & 0.65 & 0.65 & 0.65 & 50\% & 0.64 & 0.64 & 0.64 & 0.64\\
100\% & 0.62 & 0.57 & 0.58 & 0.57 & 100\% & 0.54 & 0.54 & 0.54 & 0.54\\
\hline
\end{tabular}
\end{center}
\caption{\label{beha} Behavioral Tests}
\end{table*}



Although the classification metrics analyzed in the previous section are generally regarded sufficient in estimating the performance of Bert-based models, a recent inclination of NLP researchers to behavioral testing inspired us to stress test our models as well.

For all our tests, we used our proposed RoBERTa model and applied these tests to inputs that were either titles or posts. Since we hope to extend our model to other social media platforms, we do not always expect input texts to have a title as well as a descriptive text/post. We adopted the Checklist approach \citet{beha:2020} which involve tests conducted to comprehensively analyze the model's performance.

\subsection{Synonym Replacement}
Synonym replacement is a kind of Invariance Test where label-preserving perturbations are made to the test set. As labels, the root form of the mental illnesses was chosen- {\small\verb|depress|}, {\small\verb|ptsd|}, {\small\verb|anxiou/anxiet|}, {\small\verb|bipolar|} and {\small\verb|adhd|}. Python's NLTK package and WordNet were used for these tests.

This test is conducted such that the root words are not perturbed when modifying the test set. For each post, 10\% of the tokens were randomly selected (not including the stop words or the root words). Each token was then replaced with one of its synonyms. We used the same logic for titles. We set a max and min on the number of tokens to be selected for replacement - this was (4, 30) for posts and (1, 5) for titles. Since each token was replaced with a synonym, the class label for the samples was not changed. We did this for 10, 50 and 100 percent of the test set and observed results.

In all three cases, we expect the classification metrics to drop. For the case when 10\% of the test case was modified the drop was much lower as compared to when 100\% of the test case was modified. The results are documented in Table \ref{beha}. When comparing these results to those in Table \ref{rob-class} we find that the drop in each category is about 2-4\% for posts and 5-7\% for titles. The lower drop can be attributed to the fact that synonym replacement does not alter the semantics of the input text. Therefore the model was able to draw sufficient information from the input.

\subsection{Masking}
We also performed a Directional Expectation test on the model. This is similar to the previous test but is instead performed only on labels. The labels, as defined in the previous subsection, are a list of the root form of mental illness class labels. We noticed that the root words appear often in our input texts. This behavioral test was performed to observe our model's dependency on these words. For all the tests below, we modified only those tokens that contained a root word.

In the first case, for every post from a subreddit related to a mental illness, the root form of its class label was removed from the input. For example, the input text: \emph{I feel happy for some time and then depressed again. I'm definitely bipolar} from the {\small\verb|r/bipolar|} subreddit, was modified to \emph{I feel happy for some time and then depressed again. I'm definitely}. Note that changes were not made to the word \emph{depressed} in the input. The class label for each modified sample was not changed after the perturbations. Like the previous subsection, these tests were performed on 10, 50 and 100\% of the test set.


In the second case, instead of entirely removing the tokens, we replaced it with a generic token {\small\verb|illness|}. We expected this modification to retain some semantic information that was lost in the previous test. However, we found that adding a generic token introduced some noise which reduced the overall performance of the model.

Lastly, the tokens were replaced by a randomly chosen root form of a mental illness other than its class label. With this test, we expect to force the model to pick between the label and non label tokens during classification. We believe that this is an interesting scenario to observe.

In all three cases (Table \ref{beha}), the model performance drops by some degree when compared to Table \ref{rob-class}. The first two cases showed a somewhat similar performance drop. However, the model performance was worse than that of the Synonym Replacement test. This means that the model depends on the existence of the root words in the input text to some degree.

In the third scenario, we note that the performance drop is higher. Although the test is meant to confuse the model, we observed that in some cases (especially for input: posts), we got an F1 of 0.82 with 10\% of the modified test and 0.71 with 50\% of the modified test. This is only possible if the model gathered sufficient information from the non label text in the input.


\section{Conclusion and Future Work}
Our chief motivation behind this work is the current worldwide pandemic and the mandatory confinement in many countries. We believe that social media has become the prime mode of communication for many people and has paved way for a lot of users to vent freely without judgement.

In the future we hope to be able to involve domain experts in our research and have them annotate some of our data to validate our model's performance. In addition to this, we would like to build a multi-label classifier because users may suffer from multiple mental illnesses at the same time. We would also like to work on bettering our model on the behavioral tests. Our work involves two kinds of texts- long and short - both of which are common to the internet community. Hence, our work can easily be extended to many websites. It would be interesting to collect user data from other forums and observe our model's predictions.

In conclusion, we believe that our work explores an interesting line of research where NLP is used to bridge the gap between virtual and real life of users and help those in need of medical attention.

\bibliography{naaclhlt2019}
\bibliographystyle{acl_natbib}



\end{document}